\documentclass[11pt]{article}
\usepackage[T1]{fontenc}
\usepackage[utf8]{inputenc}
\usepackage{pbox}

\usepackage{tikz}
\usetikzlibrary{shapes}
\usepackage{pgfplots}
\pgfplotsset{width=10cm,compat=1.9}

\usepgfplotslibrary{external}
\usepackage{tikz}
\usetikzlibrary{patterns,decorations.pathreplacing,calc,fadings,calc,shapes.callouts,shapes.arrows,backgrounds,spy,shadows,decorations.markings}
\immediate\write18{mkdir -p tikz}

\usetikzlibrary{intersections}
\usetikzlibrary{positioning}
\usetikzlibrary{decorations.pathreplacing}
\usetikzlibrary{matrix}
\usetikzlibrary{calc}
\usetikzlibrary{arrows,arrows.meta}
\usetikzlibrary{arrows,shapes,trees,backgrounds,shadows}
\usetikzlibrary{decorations.pathreplacing}
\usetikzlibrary{decorations.pathmorphing} % noisy shapes
\usetikzlibrary{backgrounds,snakes}
\usetikzlibrary{decorations.markings}
\usetikzlibrary{positioning}
\usetikzlibrary{plotmarks}
\usetikzlibrary{trees}
\usetikzlibrary{patterns}
\usetikzlibrary{spy}
\usetikzlibrary{calc,matrix}

\usepackage{scalerel}

\usepackage{graphicx} % Required for inserting images
\usepackage[margin=1in]{geometry}
\usepackage{amsmath}
\usepackage{caption}
\usepackage{hyperref}
\usepackage{subcaption}
\usepackage{xcolor}
\usepackage{amssymb}
\usepackage[numbers]{natbib}
\usepackage{multirow}
\usepackage{soul}

\usepackage{amsmath, amsfonts, dsfont}
\usepackage{amsthm}
\usepackage{tikz}
\usepackage{subcaption}
\usepackage{amsfonts}       % blackboard math symbols
\usepackage{amsmath}
\usepackage{amssymb}
\usepackage{booktabs}
\usepackage{longtable}
\usepackage{accents}
\usepackage{bbm}
\usepackage{hyperref}
\usepackage{placeins}
\usepackage[vlined, ruled]{algorithm2e}

\SetKwInput{KwInputs}{Inputs}
\SetKwInput{KwOutputs}{Outputs}
\SetKwRepeat{Repeat}{repeat}{until}

\newcommand{\E}{\mathbb{E}}
\newcommand{\R}{\mathbb{R}}

\usepackage{xcolor}

\begin{document}
%%%%%%%%%%%%%%%%

%\setcounter{page}{1}
%\DOI{10.1287/xxxx.0000.0000}%


%\RUNTITLE{}

% \TITLE{Bundling Information Goods of Decreasing Value}
\title{Deep Learning Method for Stationary Distribution of Reflected Brownian Motion}

% \ARTICLEAUTHORS{%
% \AUTHOR{Zhanhao Zhang}
% \AFF{Operations Research and Information Engineering, Cornell University, \EMAIL{zz564@cornell.edu}}
% \AUTHOR{Ruifan Yang}
% \AFF{Operations Research and Information Engineering, Cornell University, \EMAIL{ry298@cornell.edu}}
% \AUTHOR{Manxi Wu}
% \AFF{Operations Research and Information Engineering, Cornell University, \EMAIL{manxiwu@cornell.edu}}
% } % end of the block

\author{%
    Jim Dai\\
    \footnotesize{Operations Research and Information Engineering, Cornell University, jd694@cornell.edu}\and
    Zhanhao Zhang\\
    \footnotesize{Operations Research and Information Engineering, Cornell University, zz564@cornell.edu}
}

\date{}
\maketitle

\begin{abstract}
    The stationary distribution of reflected Brownian motion (RBM) plays an important role in the analysis of high-dimensional stochastic systems, yet closed-form solutions are known only for a few special cases. Computing important performance metrics, such as tail probabilities, is even more intractable, despite their practical relevance. In this paper, we develop a deep learning approach that accurately and efficiently learns the Laplace transform of high-dimensional RBMs based on the basic adjoint relationship (BAR). Our framework combines a careful design of the loss function, training data sampling procedure, and neural network architecture. We evaluate the proposed method on RBM instances with known ground-truth tail probabilities and demonstrate near-perfect prediction in high-dimensional settings, highlighting its potential as a general tool for analyzing stochastic systems beyond analytically tractable regimes. Our code can be found at \url{https://github.com/zhangz73/NN4MGF}.
\end{abstract}

% \KEYWORDS{High occupancy toll lane design, Non-atomic games, Equilibrium analysis with heterogeneous preferences}
% \HISTORY{}

%%%%%%%%%%%%%%%%%%%%%%%%%%%%%%%%%%%%%%%%%%%%%%%%%%%%%%%%%%%%%%%%%%%%%%

% Samples of sectioning (and labeling) in TRSC
% NOTE: (1) \section and \subsection do NOT end with a period
%       (2) \subsubsection and lower need end punctuation
%       (3) capitalization is as shown (title style).
%
%\section{Introduction.}\label{intro} %%1.
%\subsection{Duality and the Classical EOQ Problem.}\label{class-EOQ} %% 1.1.
%\subsection{Outline.}\label{outline1} %% 1.2.
%\subsubsection{Cyclic Schedules for the General Deterministic SMDP.}
%  \label{cyclic-schedules} %% 1.2.1
%\section{Problem Description.}\label{problemdescription} %% 2.

% Text of your paper here

% Text of your paper here
\section{Introduction} \label{sec:Intro}
Reflected Brownian motion (RBM) plays an important role in the analysis of multiclass queueing networks, where it often arises as a diffusion approximation under heavy traffic. In such settings, the stationary distribution of an RBM provides useful approximations for the steady-state behavior of the underlying queueing network. Closed-form expressions for stationary distributions are known only for a few special classes of RBMs. This paper develops a scalable deep learning method for computing the Laplace transform of the stationary distribution of a high dimensional RBM. The computed Laplace transforms can be used to estimate tail probabilities that serve as important performance metrics, such as tail latency for queueing networks.

We consider a $d$-dimensional RBM $Z=\{Z(t), t\ge 0\}$ associated with data $(\Sigma, \mu, R)$, which satisfies the following equations:
\begin{align*}
    &Z(t) = Z(0) + X(t) + RY(t), \qquad t \geq 0,\\
    &X = \{X(t), t \geq 0\} \text{ is a $d$-dimensional Brownian motion with} \\
    & \quad \text{covariance matrix  $\Sigma$ and drift $\mu$,} \\
    &Y(0) = 0, \quad Y(\cdot) \text{ is non-decreasing},\\
    &\int_0^{\infty} Z_k(t) dY_k(t) = 0, \qquad k = 1, \dots, d.
\end{align*}
The $d\times d$ matrix $R$ is known as the reflection matrix.
We assume that the RBM is well defined and has a unique stationary distribution $\pi$, which is satisfied, e.g., when $R$ is an $M$-matrix and $R^{-1}\mu<0$ (Harrison-Williams 1987). Define the Laplace transform of $Z$ at steady state as
\begin{align*}
    &\varphi_0(\theta) = \E_{\pi}\left[e^{\langle -\theta, Z(0) \rangle} \right], && \theta\in \R^d_+\\
    &\varphi_k(\theta) = \E_{\pi}\left[\int_0^1 e^{\langle -\theta, Z(t) \rangle} dY_k(t) \right], && k = 1, \dots, d,
\end{align*}
where $Z(0)$ follows the stationary distribution $\pi$. Lemma 1 of \cite{dai2014multi} shows that the Laplace transforms $\varphi$ and $\varphi_k$ are uniquely characterized by the Laplace version of basic adjoint relationship (BAR)
\begin{align} \label{eq:BAR}
    \gamma_0(\theta) \varphi_0(\theta) =& \sum_{k = 1}^d \gamma_k(\theta) \varphi_k(\theta),
\end{align}
where $\gamma_0(\theta) = -\frac{1}{2} \langle \theta, \Sigma \theta \rangle + \langle \mu, \theta \rangle$ and $\gamma_k(\theta) = -\langle R^{(k)}, \theta \rangle$. Here, $R^{(k)}$ denotes the $k$-th column of $R$. In \cite{dai2014multi}, the authors prove that the Laplace version BAR (\ref{eq:BAR}) is equivalent to the PDE version of the BAR that was first advanced in Harrison-Williams (1987).

In contrast to \cite{blanchet2021efficient}, which develops an efficient approach for estimating steady-state expectations of high-dimensional RBMs, our work targets the Laplace transforms of their stationary distributions. Since the Laplace transform can be numerically inverted to recover tail probabilities and, more broadly, the full stationary distribution \citep{abate2006unified}, it provides a significantly richer characterization of system performance. To the best of our knowledge, this is the first framework for estimating the Laplace transform $\varphi_k(\cdot)$ of high-dimensional RBMs. The main methodological contribution is to turn the BAR characterization into a scalable learning problem by combining a carefully designed loss function tailored to the BAR and the structural properties of Laplace transforms, a targeted training data sampling scheme, and a neural network architecture whose number of trainable parameters does not scale with the dimension $d$. This yields an accurate and efficient computational tool for evaluating important performance metrics, such as tail probabilities and tail latencies, in large-scale stochastic systems beyond analytically tractable regimes.
%We develop a deep learning approach for this problem, with the potential to provide practitioners an accurate and efficient tool for evaluating important performance metrics, such as tail probabilities and tail latencies, in large-scale stochastic systems beyond analytically tractable regimes.

\paragraph{Literature review.} 
%% Numerical approaches in queueing
With the advancement of computational power, there has been growing interest in developing numerical methods for stochastic systems. For instance, \cite{qu2024deep} propose a deep learning approach to compute convergence rates of Markov chains, and \cite{qu2026deep} extend this framework to estimate Lyapunov functions, solve Poisson equations, and approximate stationary distributions. However, these methods are currently limited to low-dimensional settings (e.g., two dimensions). Extending deep learning methods to high-dimensional stochastic systems remains significantly more challenging and has motivated a growing line of research in stochastic control and diffusion approximations.

%% Deep BSDE and stochastic systems.
Our work is closely related to this literature on applying deep learning to high-dimensional stochastic systems. In particular, the deep BSDE framework \citep{han2017deep, han2018solving} provides a powerful approach for solving high-dimensional stochastic control problems by leveraging connections between parabolic PDEs and backward stochastic differential equations. This framework has been successfully applied to a wide range of settings, including matching \citep{ata2025dynamic}, scheduling \citep{ata2025scheduling}, impulse control \citep{ata2025computational}, drift control \citep{ata2025drift}, and singular control \citep{ata2024singular}. Our work is also motivated by stochastic systems, as the equation \eqref{eq:BAR} characterizes the stationary behavior of reflected Brownian motions. However, our setting is fundamentally different: unlike these approaches, which often exploit stochastic representations, such as the Feynman--Kac representation, to reformulate high-dimensional PDEs as stochastic differential equations, \eqref{eq:BAR} describes a stationary relationship and does not admit such a reformulation.

%% Least squares for variational problems
Our work is also related to the extensive literature on applying deep learning to high-dimensional partial differential equations (PDEs); see the survey by \cite{weinan2021algorithms}. In particular, prior work has studied PDEs arising from variational formulations using deep learning \citep{carleo2017solving, yu2018deep, sirignano2018dgm}. Our approach constructs a loss function based on the squared residual of \eqref{eq:BAR}, which parallels the least-squares formulations used in \cite{carleo2017solving, sirignano2018dgm}. However, directly minimizing this residual is insufficient in our setting due to generalizability and numerical stability issues. To address this, we design a structured loss function with additional regularization terms that enforce key properties of the Laplace transform. In addition, we propose a tailored sampling scheme and a neural network architecture that scale effectively to high-dimensional problems.

\section{Deep learning approach} \label{sec:DL-approach}

In this section, we approximate $\varphi_k(\cdot)$ for $k = 0, \dots, d$ in \eqref{eq:BAR} using feedforward neural networks in the complex domain, where \eqref{eq:BAR} continues to hold by analytic extension. Working in the complex domain is necessary because robust numerical inversion methods, such as the Talbot method \citep{talbot1979accurate} and related approaches \citep{weideman2006optimizing, abate1992fourier}, evaluate the Laplace transform at complex arguments when computing tail probabilities. Let $\theta=\theta^{\mathrm{Re}}+i\theta^{\mathrm{Im}} \in \mathbb{C}^d$ be a $d$-dimensional complex vector, with $\theta^{\mathrm{Re}},\theta^{\mathrm{Im}}\in\mathbb{R}^d$. We focus on accurate approximation over the bounded region
\begin{align*}
    \Theta := [\underline{\theta}^{\mathrm{Re}}, \bar{\theta}^{\mathrm{Re}}]^d \times [-\bar{\theta}^{\mathrm{Im}}, \bar{\theta}^{\mathrm{Im}}]^d.
\end{align*}
A naive approach is to use shallow feedforward neural networks to approximate the functions $\varphi_k(\cdot)$, and then update the network parameters by minimizing the mean-squared error between the left-hand side and the right-hand side of \eqref{eq:BAR}, where the training samples are drawn uniformly from $\Theta$. This straightforward framework, however, suffers from several fundamental difficulties.

\paragraph{Poor generalization.} The equation \eqref{eq:BAR} characterizes the Laplace transform of the stationary distribution, which is analytic and monotone along the real axis. However, the naive training framework only minimizes a finite-sample residual of \eqref{eq:BAR} and does not enforce either analyticity or monotonicity. As a result, the neural network may fit the sampled training points well while violating these structural properties and deviating significantly at unseen inputs. 

\paragraph{Numerical stability.} Because the Laplace transforms $\varphi_k(\theta)$ vary exponentially with $\theta$, their values can differ by many orders of magnitude within $[\underline{\theta}, \bar{\theta}]^d$, leading to numerical precision issues when $\theta$ is either small or large. 

\paragraph{Imbalanced training data.} As $d$ grows, samples drawn uniformly from high-dimensional boxes for the real and imaginary parts of $\theta$ rarely fall in corner regions, where many coordinates are simultaneously close to their extreme values. Instead, a typical sample contains a mixture of small and large coordinate values across dimensions. Consequently, the naive sampling scheme under-represents regimes in which the Laplace transform exhibits large magnitudes or strong oscillatory behavior, making it difficult for the training algorithm to learn these regions accurately. 

\paragraph{Poor scalability.} A standard feedforward network takes the full vector $\theta \in \mathbb{C}^d$ as input, so the size of its input layer grows linearly with $d$. As a result, the first layer contains parameters tied to individual input coordinates, which must be learned from samples that capture sufficient variability of $\theta$ across all $d$ dimensions. As $d$ increases, achieving such coverage requires substantially larger sample sizes, making the naive architecture less scalable for high-dimensional RBMs.

In the rest of this section, we will describe how we address these issues through a careful design of the loss function, training data sampling, and neural network architecture.

\subsection{Loss function}
Instead of directly learning the Laplace transforms $\varphi_k(\cdot)$, we parameterize their logarithms using neural networks. Specifically, the networks output functions $f_k(\cdot)$ such that
\begin{align*}
    \varphi_k(\theta) = \exp(f_k(\theta)), \qquad k = 0, \dots, d.
\end{align*}
This log-parameterization improves numerical stability and allows us to work with quantities of comparable scale when evaluating the BAR equation.

For any $\theta \in \Theta$, we define the loss function
\begin{align}
    \mathcal{L}(\theta) :=& \mathcal{L}_{{\rm BAR}}(\theta) + \lambda_{\rm pair} \cdot \mathcal{L}_{\rm pair}(\theta) + \lambda_{\rm mono} \cdot \mathcal{L}_{\rm mono}(\theta) \nonumber \\
    +& \lambda_{\rm CR} \cdot \mathcal{L}_{\rm CR}(\theta) + \lambda_{\rm zero} \cdot \mathcal{L}_{\rm zero}, \label{eq:loss}
\end{align}
where $\lambda_{\rm pair}, \lambda_{\rm mono}, \lambda_{\rm CR}, \lambda_{\rm zero} > 0$ are hyperparameters that balance the relative importance of the different loss components. The first term enforces the BAR equation, while the remaining terms incorporate structural properties that the Laplace transforms are known to satisfy.

\subsubsection{Normalized BAR error.}
The primary objective is to enforce the BAR equation \eqref{eq:BAR}. However, directly comparing the two sides of the equation can lead to numerical instability because the Laplace transforms may vary exponentially with $\theta$. To mitigate this issue, we introduce a normalization factor.

For any input $\theta \in \Theta$, we first compute the normalization factor $\nu(\theta)$ as:
\begin{align*}
    \nu(\theta) :=& \max_{k = 0, \dots, d} \left\{ \log \vert \gamma_k(\theta) \vert + f^{Re}_{k}(\theta) \right\}.
\end{align*}
This normalization rescales the two sides of the BAR equation to a comparable magnitude before evaluating their difference. We then compute
\begin{align*}
    \kappa_{l}(\theta) =& \gamma_0(\theta) \cdot \exp\Big(f_{0}(\theta) - \nu(\theta) \Big),\\
    \kappa_{r}(\theta) =& \sum_{k = 1}^d \gamma_k(\theta) \cdot \exp\Big(f_{k}(\theta) - \nu(\theta) \Big).
\end{align*}
Finally, the normalized BAR error is given as
\begin{align}
    \mathcal{L}_{\rm BAR}(\theta) :=& \left(\frac{\vert \kappa_l(\theta) - \kappa_r(\theta) \vert}{\vert \kappa_l(\theta) \vert + \vert \kappa_r(\theta) \vert + \epsilon} \right)^2, \label{eq:loss-bar}
\end{align}
where $\epsilon > 0$ is a very small constant to avoid division by zero. This normalized error stabilizes training by comparing the relative discrepancy between the two sides of \eqref{eq:BAR}.

\subsubsection{Pairwise consistency penalty.}
Next, we exploit structural constraints implied by the BAR equation by constructing points at which only one boundary term remains active. Given input $\theta \in \Theta$, we construct $\tilde{\theta}^{(1)}, \dots, \tilde{\theta}^{(d)}$ that satisfies $R_{-k} \tilde{\theta}^{(k)} = 0$ and $\tilde{\theta}^{(k)}_k = \theta_k$, where $R_{-k}$ denotes the matrix obtained from $R$ by removing its $k$-th row. By construction, these vectors lie on the boundary where only one reflection term remains active. In particular, for any $k, k' = 1, \dots, d$ such that $k \neq k'$, we have $\gamma_{k'}(\tilde{\theta}^{(k)}) = 0$.

The original BAR equation relates the interior function only to the aggregate contribution of all boundary functions. Consequently, each boundary function is constrained only indirectly through this aggregate residual, making it difficult to learn the individual boundary functions efficiently. We therefore use the constructed points above to directly relate each boundary function to the interior function, providing an explicit constraint for each individual boundary function. Substituting this relation into \eqref{eq:BAR} and taking logarithms yields the following pairwise consistency penalty, which substantially accelerates convergence:
\begin{align}
    &\mathcal{L}_{\rm pair}(\theta) 
    := \sum_{k = 1}^d \Big(\log \gamma_0(\tilde{\theta}^{(k)}) + f_{0}(\tilde{\theta}^{(k)}) - \log \gamma_k(\tilde{\theta}^{(k)}) - f_{k}(\tilde{\theta}^{(k)}) \Big)^2. \label{eq:loss-pairwise-consistency}
\end{align}

\subsubsection{Monotonicity penalty.}
The Laplace transforms are non-increasing in the real part of $\theta$. To enforce this property in the learned functions, we introduce a monotonicity penalty. Given any input $\theta \in \Theta$, we construct $\tilde{\theta} := Re(\theta) + 0i$, which lies on the real axis.
\begin{align}
    &\mathcal{L}_{\rm mono}(\theta) 
    := \sum_{k = 0}^d \left(\frac{\partial f^{\rm Re}_{k}(\tilde{\theta})}{\partial \rm Re(\tilde{\theta})} \right)^+ + \lambda_{\rm img} \sum_{k = 0}^d (f^{\rm Im}_k(\tilde{\theta}))^2, \label{eq:loss-mono}
\end{align}      
where $\lambda_{\rm img} > 0$ is a hyperparameter. The first term penalizes violations of monotonicity, while the second enforces real-valued outputs when $\theta$ lies on the real axis.

\subsubsection{Cauchy-Riemann penalty.}
Since the Laplace transforms are analytic functions, their real and imaginary parts must satisfy the Cauchy–Riemann equations. We therefore include the following penalty to enforce approximate analyticity:
\begin{align}
    &\mathcal{L}_{\rm CR}(\theta) 
    := \sum_{k = 0}^d \left[\left(\frac{\partial f^{\rm Re}_{k}(\theta)}{\partial \rm Re(\theta)} - \frac{\partial f^{\rm Im}_{k}(\theta)}{\partial \rm Im(\theta)} \right)^2 + \left(\frac{\partial f^{\rm Re}_{k}(\theta)}{\partial \rm Im(\theta)} + \frac{\partial f^{\rm Im}_{k}(\theta)}{\partial \rm Re(\theta)} \right)^2 \right]. \label{eq:loss-CR}
\end{align}

\subsubsection{Zero-anchoring penalty.}
Finally, we anchor the interior transform at the origin. Since $\varphi_0(0)=1$, we enforce this normalization through
\begin{align}
    \mathcal{L}_{\rm zero} := \left\Vert\exp(f_{0}(0)) - 1\right\Vert_2^2. \label{eq:loss-zero}
\end{align}

\subsection{Sampling strategy}
At each gradient update for minimizing the loss function \eqref{eq:loss}, we sample a batch of $N$ data points from $\Theta$ to compute the loss and its gradient with respect to the neural network parameters. As discussed above, uniform sampling becomes increasingly ineffective in high dimensions because it rarely samples targeted corner regions. These regions are important for capturing large-magnitude Laplace values in the real domain and strong oscillatory behavior in the imaginary domain, both of which can significantly affect the accuracy of numerical inversion methods such as the Talbot method. To improve coverage of these regions while maintaining support over the full domain $\Theta$, we adopt the following two-stage sampling procedure.

We sample each training data point $\theta \in \Theta$ in a batch as follows:
\begin{enumerate}
    \item \textbf{Step 1:} We first sample scalar reference thresholds
    $\theta^{\mathrm{Re}}_* \sim {\rm Unif}[\underline{\theta}^{\mathrm{Re}}, \bar{\theta}^{\mathrm{Re}}]$ and $\theta^{\mathrm{Im}}_* \sim {\rm Unif}[0, \bar{\theta}^{\mathrm{Im}}]$. These thresholds determine the targeted corner regions for the real and imaginary parts.
    \item \textbf{Step 2:} Conditional on these thresholds, we sample $\theta^{\mathrm{Re}} \sim {\rm Unif}[\underline{\theta}^{\mathrm{Re}}, \theta^{\mathrm{Re}}_*]^d$ and $\theta^{\mathrm{Im}} \sim {\rm Unif}\big[([-\bar{\theta}^{\mathrm{Im}}, -\theta^{\mathrm{Im}}_*] \cup [\theta^{\mathrm{Im}}_*, \bar{\theta}^{\mathrm{Im}}])^d\big]$, and construct $\theta = \theta^{\mathrm{Re}} + i\theta^{\mathrm{Im}}$. This conditional sampling scheme increases the probability of drawing points near the lower corner of the real domain and near the outer corners of the imaginary domain.
\end{enumerate}
We use this procedure to generate all training samples in each batch, with the batch size $N$ chosen independently of the dimension $d$.

\subsection{Neural network architecture}
\begin{figure}[h]
    \centering
    \begin{subfigure}{0.45\linewidth}
        \centering
        \includegraphics[width=\linewidth]{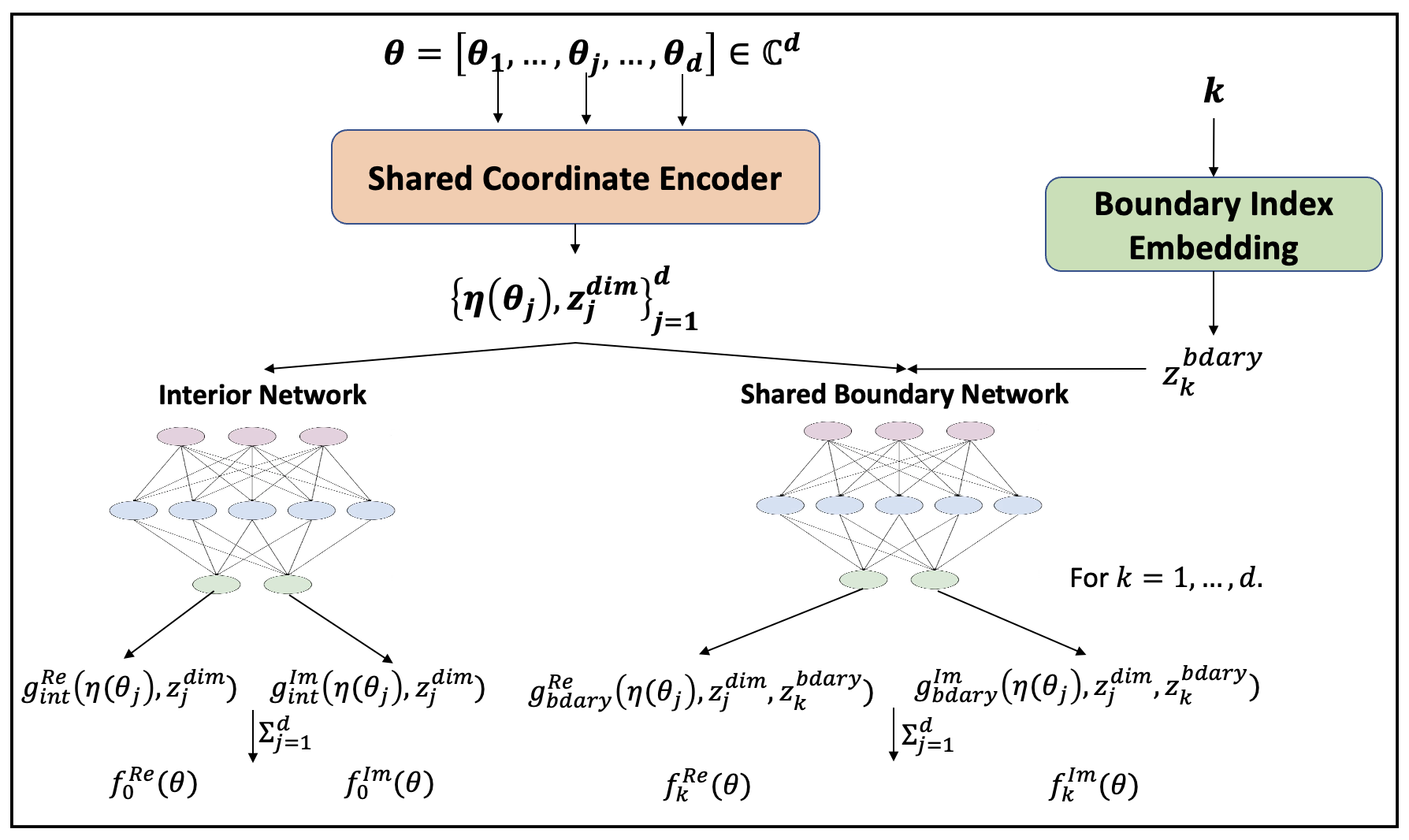}
        \caption{Full neural network pipeline}
        \label{fig:nn-pipeline}
    \end{subfigure}%
    \begin{subfigure}{0.45\linewidth}
        \centering
        \includegraphics[width=\linewidth]{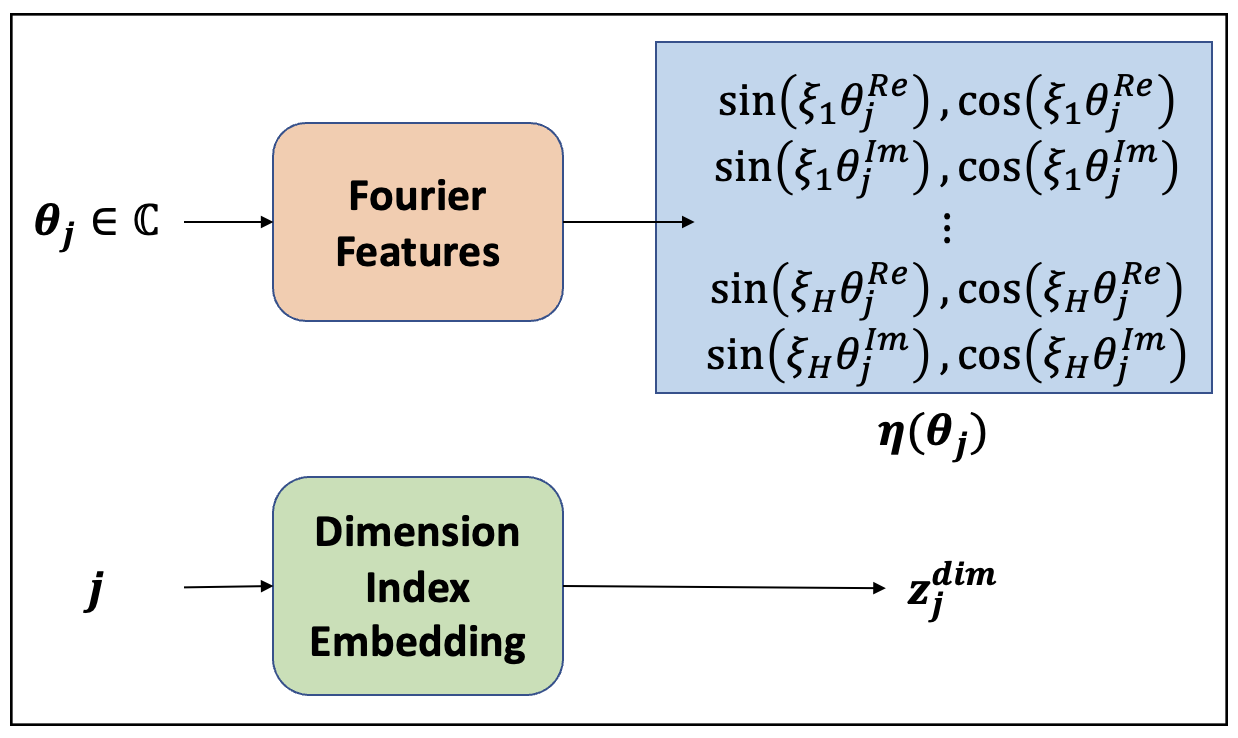}
        \caption{Shared coordinate encoder}
        \label{fig:nn-coordinate-encoder}
    \end{subfigure}
    \caption{Neural network architecture}
    \label{fig:nn}
\end{figure}

To address the scalability challenges discussed earlier, we design a neural network architecture whose size does not grow with the dimension $d$, which encodes each coordinate using a shared feature extractor and aggregates the resulting representations to produce the interior and boundary functions (see Figure \ref{fig:nn-pipeline}).

\paragraph{Shared coordinate encoder.}
Given $\theta = [\theta_1,\dots,\theta_d] \in \Theta$, we first apply a shared encoder to each coordinate $\theta_j$, which combines two components (see Figure~\ref{fig:nn-coordinate-encoder}). The first component computes Fourier features $\eta(\theta_j)$ of the real and imaginary parts of $\theta_j$, which improves the ability of the network to represent oscillatory patterns. The second component maps the coordinate index $j$ to an embedding vector $z_j^{\rm dim}$, allowing the network to distinguish different coordinates while keeping the encoder weights shared across all dimensions. 

\paragraph{Interior network.}
For each coordinate $j = 1, \dots, d$, the corresponding Fourier features $\eta(\theta_j)$ and the coordinate embedding vector $z_j^{\rm dim}$ are then passed to an interior network (i.e. a feedforward neural network with two heads) $g_{\rm int}$. The network outputs a two-dimensional vector $g_{\rm int}\big(\eta(\theta_j), z_j^{\rm dim}\big) =
\big(g_{\rm int}^{\rm Re}\big(\eta(\theta_j), z_j^{\rm dim}\big),\, g_{\rm int}^{\rm Im}\big(\eta(\theta_j), z_j^{\rm dim}\big)\big)$, where $g_{\rm int}^{\rm Re}\big(\eta(\theta_j), z_j^{\rm dim}\big)$ (resp. $g_{\rm int}^{\rm Im}\big(\eta(\theta_j), z_j^{\rm dim}\big)$) represents the real (resp. imaginary) contribution of coordinate $j$ to the log of the interior Laplace transform. These contributions are aggregated across coordinates to obtain $f^{\rm Re}_{0}(\theta) = \sum_{j=1}^d g_{\rm int}^{\rm Re}\big(\eta(\theta_j), z_j^{\rm dim}\big)$ and $f^{\rm Im}_{0}(\theta) = \sum_{j=1}^d g_{\rm int}^{\rm Im}\big(\eta(\theta_j), z_j^{\rm dim}\big)$. This additive structure allows the network to handle high-dimensional inputs while maintaining a parameter count that does not scale with $d$.

\paragraph{Shared boundary network.}
To represent the boundary functions, we use another feedforward network $g_{\rm bdary}$ with shared weights across all boundaries. For each boundary $k = 1, \dots, d$ and coordinate $j = 1, \dots, d$, in addition to the Fourier features $\eta(\theta_j)$ and the coordinate embedding vector $z_j^{\rm dim}$, the network also receives a boundary index embedding $z_k^{\rm bdary}$. The network outputs coordinate-wise contributions $g_{\rm bdary}\big(\eta(\theta_j), z_j^{\rm dim}, z_k^{\rm bdary} \big)$, which are again aggregated to produce $f_k^{\rm Re}(\theta) = \sum_{j=1}^d g_{\rm bdary}^{\rm Re}\big(\eta(\theta_j), z_j^{\rm dim}, z_k^{\rm bdary}\big)$ and $f_k^{\rm Im}(\theta) = \sum_{j=1}^d g_{\rm bdary}^{\rm Im}\big(\eta(\theta_j), z_j^{\rm dim}, z_k^{\rm bdary}\big)$ for $k=1, \ldots, d$.

Overall, this architecture leverages shared encoders and additive aggregation to achieve scalability in high dimensions while allowing the network to capture both interior and boundary behaviors of the Laplace transforms.

\section{Numerical Experiments}
We evaluate the performance of our neural network by estimating tail probabilities of the form $\mathbb{P}(\sum_j Z_j > t)$. The predicted probabilities are obtained by applying numerical inverse Laplace transforms using the Talbot method \citep{talbot1979accurate} to the Laplace transform learned by the neural network. To assess the accuracy of this approach, we consider two RBM examples for which ground truth results of tail probabilities can be obtained.

The first example is a $2$-dimensional RBM from \cite{dai1992reflected} and \cite{harrison1978diffusion}, which admits a closed-form expression for the stationary density but does not provide an explicit formula for the Laplace transform. In this case, we compute the ground-truth probabilities $\mathbb{P}(\sum_j Z_j > t)$ by numerically integrating the density function. The second and third examples are $20$-dimensional and $30$-dimensional RBMs from \cite{dai2014multi}, whose Laplace transform admits a product-form expression. The corresponding ground-truth probabilities are computed by applying the same Talbot inversion method to the exact Laplace transform. Both numerical integration and Talbot inversion of Laplace transform can be achieved using the \texttt{mpmath} package in Python \citep{mpmath}.

\begin{figure}[t]
    \centering
    \begin{subfigure}{0.30\linewidth}
        \centering
        \includegraphics[width=\linewidth]{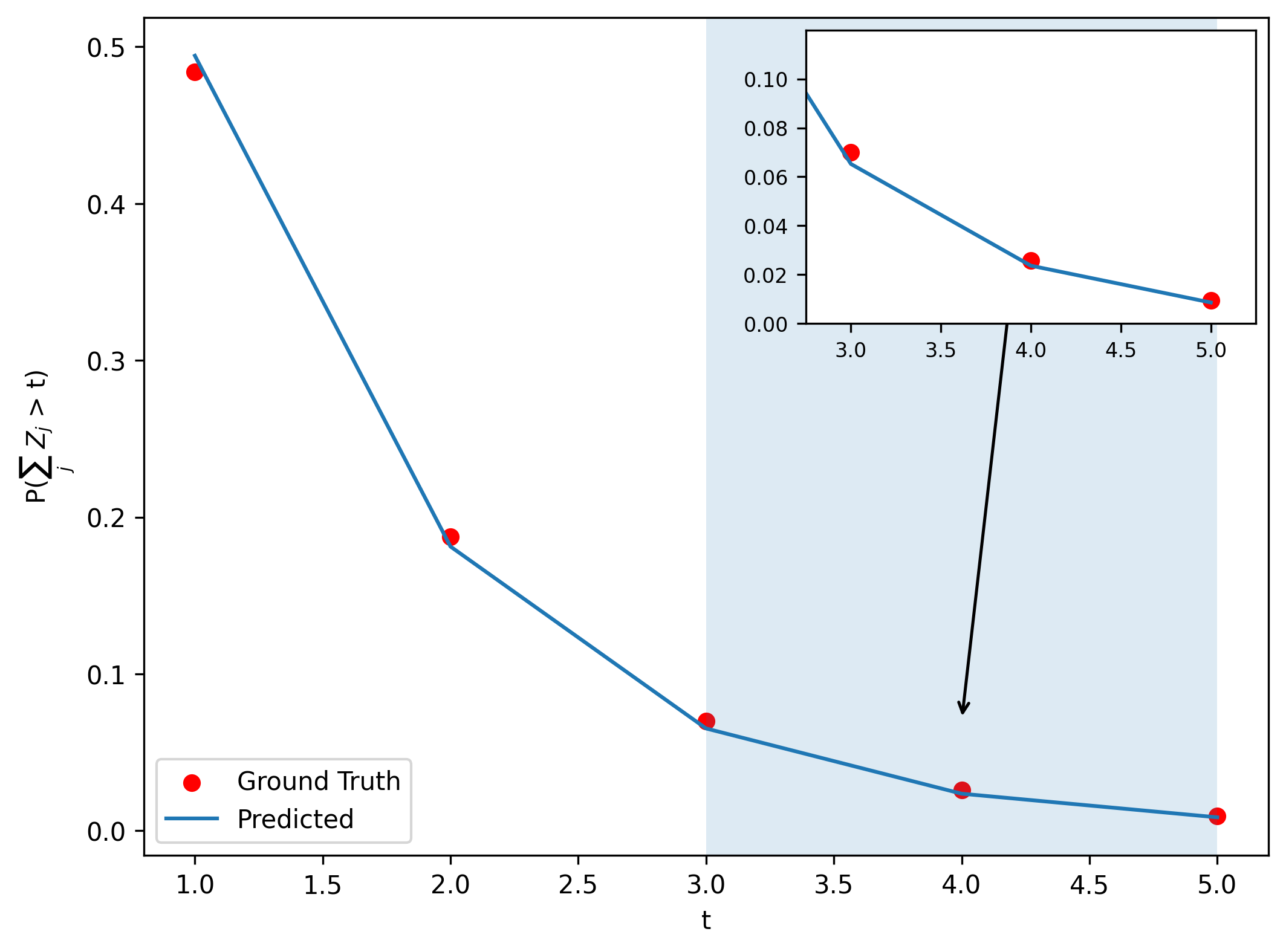}
        \caption{$2$-d RBM without closed-form Laplace transform}
    \end{subfigure}%
    \quad
    \begin{subfigure}{0.30\linewidth}
        \centering
        \includegraphics[width=\linewidth]{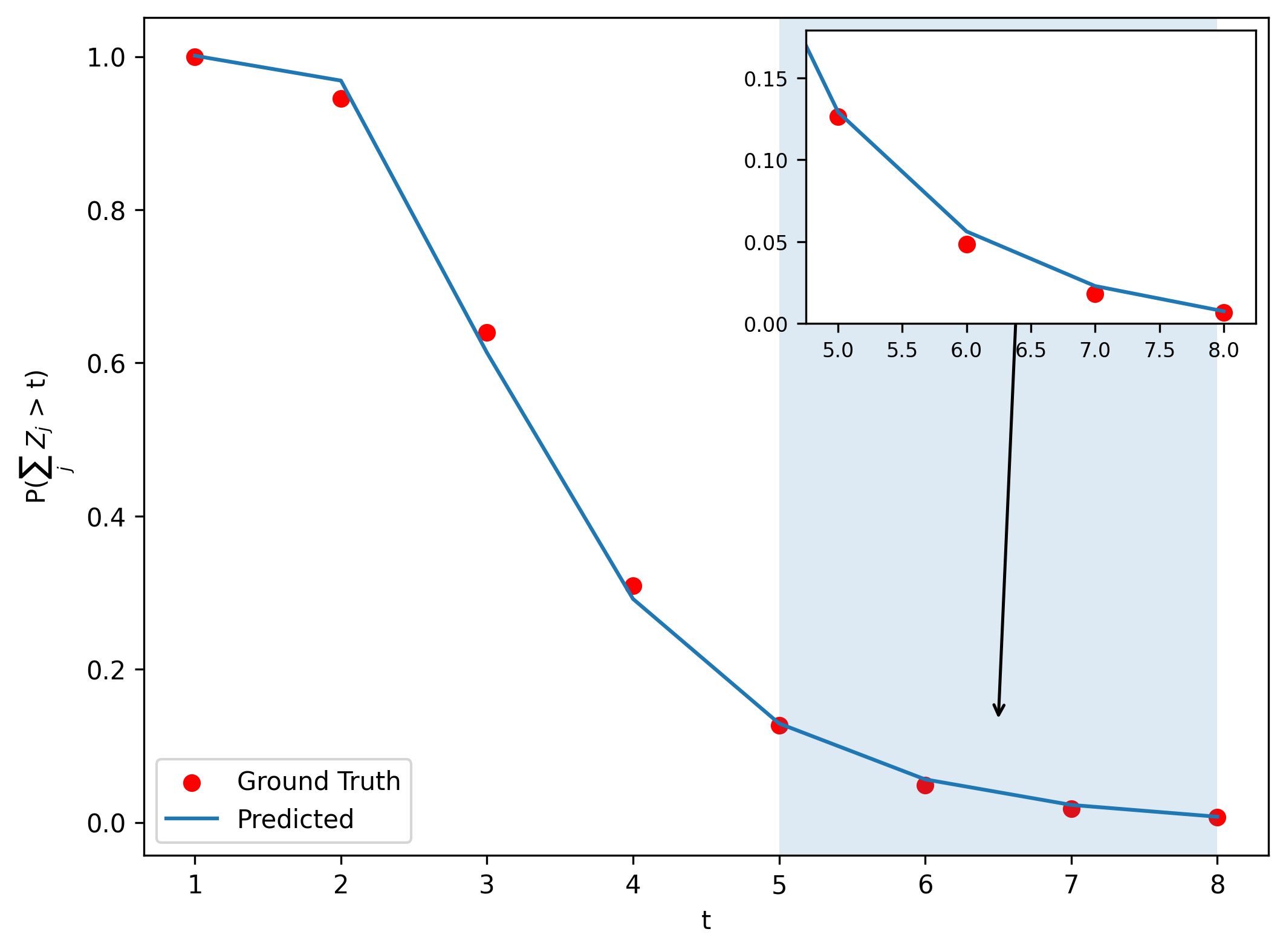}
        \caption{$20$-d RBM with product form Laplace transform}
    \end{subfigure}
    \quad
    \begin{subfigure}{0.30\linewidth}
        \centering
        \includegraphics[width=\linewidth]{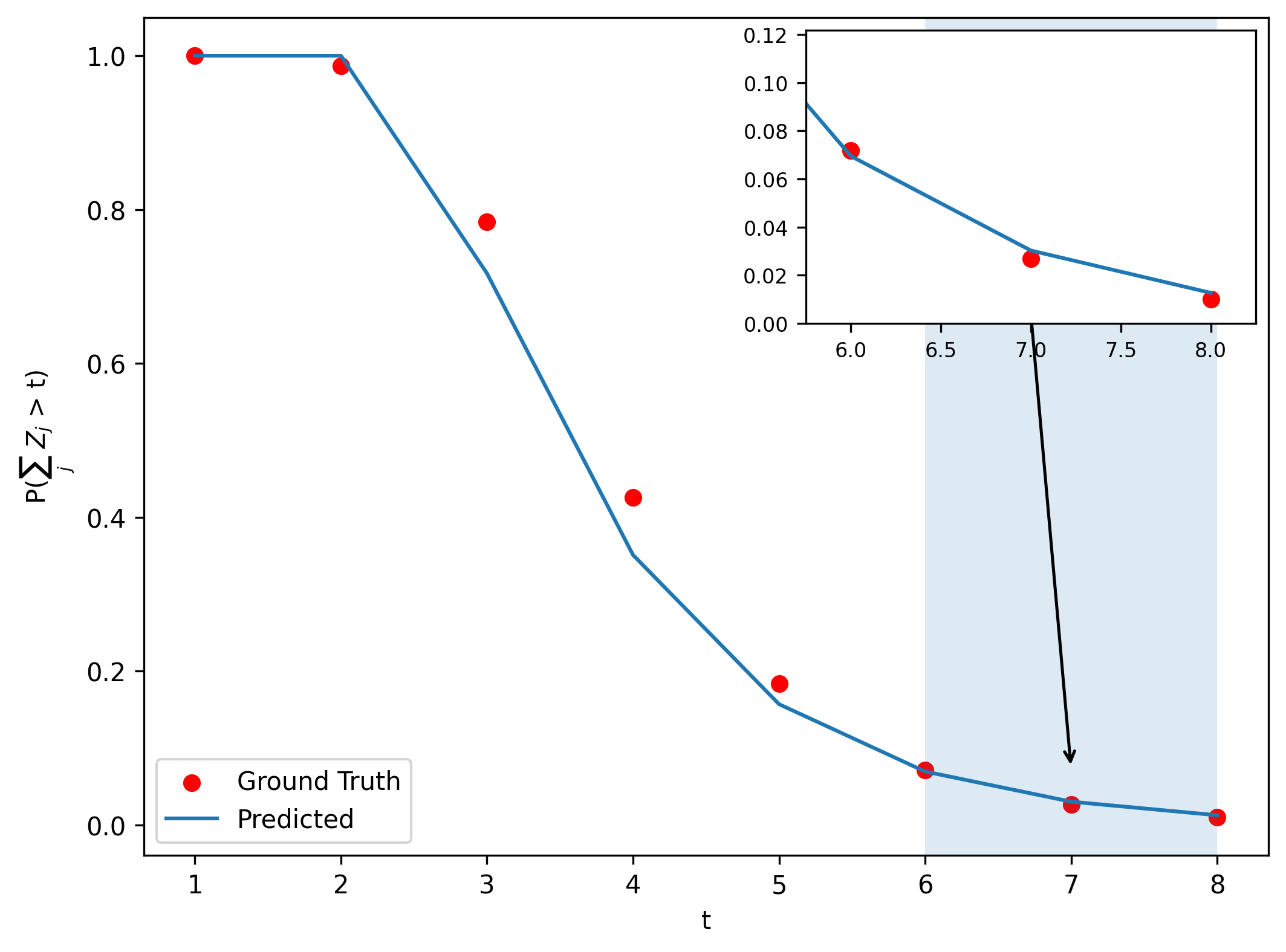}
        \caption{$30$-d RBM with product form Laplace transform}
    \end{subfigure}
    \caption{Tail probabilities (up to 1\%) of the stationary sum $\sum_j Z_j$: neural network vs.\ ground truth.}
    \label{fig:tail-probs}
\end{figure}

The predicted tail probabilities match the ground truth almost perfectly in both examples (see Figure \ref{fig:tail-probs}). The two-dimensional example shows that our neural network can capture Laplace transforms with complex structures, where the behavior differs significantly across dimensions. The $20$-dimensional and $30$-dimensional examples highlight the scalability of the architecture: despite the high dimensionality, the neural network still produces highly accurate probability estimates. These results indicate that our neural network is both expressive enough to represent complex Laplace transforms and scalable to high-dimensional RBMs. 

\section{Conclusion}
In this paper, we propose a scalable deep learning framework for learning the Laplace transform of high-dimensional RBMs. Our approach integrates a carefully designed loss function, training data sampling procedure, and neural network architecture. Numerical experiments demonstrate that, by numerically inverting the learned Laplace transform, our method can accurately compute tail probabilities of the stationary distribution in high-dimensional settings. These results highlight the potential of our framework as a general tool for estimating performance metrics in large-scale stochastic systems.

Despite these promising results, several limitations remain. First, each gradient update currently requires $16{,}384$ training samples, leading to substantial GPU memory usage when processed in a single batch, or increased runtime when split across multiple batches, as the dimension of the RBM grows. Developing more efficient training strategies to scale our approach to RBMs with hundreds or thousands of dimensions is an important direction for future work. Second, beyond RBMs, it would be of interest to extend our framework to broader classes of stochastic systems and estimate practically relevant performance metrics, such as tail latencies and throughput.

%\THEEndNotes
% \begingroup \parindent 0pt \parskip 0.0ex \def\enotesize{\normalsize} \theendnotes \endgroup

% Appendix here
% Options are (1) APPENDIX (with or without general title) or
%             (2) APPENDICES (if it has more than one unrelated sections)
% Outcomment the appropriate case if necessary
%
% \begin{APPENDIX}{<Title of the Appendix>}
% \end{APPENDIX}
%
%   or
%
\clearpage

\appendix
\section{Implementation Details}
\paragraph{Construction of Fourier features.} To enhance the representation of complex inputs, we employ Fourier feature embeddings applied separately to the real and imaginary parts. Consider an input $\theta := \theta^{\rm Re} + i\theta^{\rm Im} \in \mathbb{C}^d$. For each coordinate $j = 1, \dots, d$, we first normalize the inputs so that both the real and imaginary parts lie in $[-1,1]$. Specifically, we define the normalized real and imaginary components $\tilde{\theta}^{\rm Re}_j$ and $\tilde{\theta}^{\rm Im}_j$ as
\begin{align*}
    \tilde{\theta}^{\rm Re}_j :=& 2 \frac{\theta^{\rm Re}_j - \underline{\theta}^{\rm Re}_j}{\bar{\theta}^{\rm Re}_j- \underline{\theta}^{\rm Re}_j} - 1,\\
    \tilde{\theta}^{\rm Im}_j :=& \frac{\theta^{\rm Im}_j + \bar{\theta}^{\rm Im}_j}{\bar{\theta}^{\rm Im}_j} - 1.
\end{align*}
We then construct Fourier features based on these normalized inputs. Let $H$ denote the total number of Fourier features for each of the real and imaginary parts, where $H$ is assumed to be divisible by $2$. We construct two sets of frequency grids $\{\xi_h^{\rm Re}\}_{h=1}^{H/2}$ and $\{\xi_h^{\rm Im}\}_{h=1}^{H/2}$ as fixed, non-trainable hyperparameters, using logarithmically spaced values over $[\delta_{\min}^{\rm Re}, \delta_{\max}^{\rm Re}]$ and $[\delta_{\min}^{\rm Im}, \delta_{\max}^{\rm Im}]$, respectively. The resulting $2H$ Fourier features are given by
\begin{align*}
    \Big\{\sin\Big(2\pi \xi^{\rm Re}_h \theta_j^{\rm Re}\Big), \cos\Big(2\pi \xi^{\rm Re}_h \theta_j^{\rm Re}\Big), \sin\Big(2\pi \xi^{\rm Im}_h \theta_j^{\rm Im}\Big), \cos\Big(2\pi \xi^{\rm Im}_h \theta_j^{\rm Im}\Big)\Big\}_{h = 1, \dots, H/2}.
\end{align*}

\paragraph{Hyperparameters for neural networks.} In all experiments, we set $H = 64$, with $\delta_{\min}^{\rm Re} = \delta_{\min}^{\rm Im} = -4$, $\delta_{\max}^{\rm Re} = 1$, and $\delta_{\max}^{\rm Im} = 2$. The dimension index embedding $z_j^{\rm dim}$ and the boundary index embedding $z_k^{\rm bdary}$ are both chosen to have dimension $64$. For the interior network and the shared boundary network, we use feedforward architectures with two hidden layers, each consisting of $128$ neurons and \texttt{SiLU} activation.

\paragraph{Hyperparameters for training.} For all experiments, we use the \texttt{AdamW} optimizer with a \texttt{CosineAnnealing} learning rate schedule, both implemented in \texttt{PyTorch}. The learning rate is initialized at $10^{-3}$ and annealed to $10^{-4}$. We train the neural networks for $100{,}000$ epochs, where each epoch corresponds to a single gradient update. For the $20$- and $30$-dimensional examples, we further fine-tune the neural networks for an additional $200{,}000$ epochs. Specifically, the learning rate is annealed from $10^{-4}$ to $10^{-5}$ over the first $100{,}000$ fine-tuning epochs, and then from $10^{-5}$ to $10^{-6}$ over the remaining $100{,}000$ epochs. At each epoch, we sample $2^{14} = 16{,}384$ training data points to estimate the gradient of the loss function \eqref{eq:loss}.

To reduce memory consumption, we evaluate the more expensive loss components \eqref{eq:loss-pairwise-consistency}--\eqref{eq:loss-CR} using random subsets of these $16{,}384$ training data points. The pairwise consistency loss \eqref{eq:loss-pairwise-consistency} is evaluated using only $300$ points. This is because PyTorch stores the intermediate activations of the neural network for each input until backpropagation, and each input $\theta$ is transformed into $d$ auxiliary inputs $\tilde{\theta}^{(1)},\dots,\tilde{\theta}^{(d)}$, causing the memory requirement to grow linearly with $d$. For the monotonicity penalty \eqref{eq:loss-mono} and Cauchy--Riemann penalty \eqref{eq:loss-CR}, we first use $1024$ points to evaluate the interior terms (i.e., $k=0$), and then use $128$ points to evaluate these penalties across all $k=0,\dots,d$. These penalties are particularly memory intensive because they require first-order derivatives of the neural network outputs, which require retaining the computational graph so that gradients can be backpropagated through these derivatives. The penalty coefficients are set to $\lambda_{\rm bdary} = \lambda_{\rm mono} = \lambda_{\rm CR} = 10$ and $\lambda_{\rm zero} = 0.1$.

% To reduce memory consumption, we evaluate the more expensive loss components \eqref{eq:loss-pairwise-consistency}--\eqref{eq:loss-CR} using random subsets of these $16{,}384$ training data points. Specifically, we use $300$ points for the pairwise consistency loss \eqref{eq:loss-pairwise-consistency}. For the monotonicity and Cauchy–Riemann penalties \eqref{eq:loss-mono} and \eqref{eq:loss-CR}, we first use $1024$ points to evaluate the interior terms (i.e., $k=0$), and then use $128$ points to evaluate these penalties across all $k = 0, \dots, d$. The penalty coefficients are set to $\lambda_{\rm bdary} = \lambda_{\rm mono} = \lambda_{\rm CR} = 10$ and $\lambda_{\rm zero} = 0.1$.

\paragraph{Problem instances.} We consider two classes of RBMs for all our numerical experiments. The first is adopted from \citep{dai1992reflected}, where a closed-form expression is available for the stationary distribution, but not for its Laplace transform or tail probabilities. Specifically, we consider a $2$-dimensional RBM with
\begin{align*}
    \Sigma = \begin{bmatrix}
        1 & 0\\0 & 1\\
    \end{bmatrix}, \qquad \mu = \begin{bmatrix}
        -1 \\ 0
    \end{bmatrix}, \qquad R = \begin{bmatrix}
        1 & 0 \\ -1 & 1
    \end{bmatrix}.
\end{align*}
The density $\rho$ of its stationary distribution is given by
\begin{align*}
    \rho(\theta) =& Cr^{1/2} e^{-(r+\theta_1)} \cos(\psi/2),
\end{align*}
where $C$, $r$, and $\psi$ are defined as
\begin{align*}
    C =& \frac{1}{\sqrt{\pi}} 2^{3/2},\\
    r =& \sqrt{\theta_1^2 + \theta_2^2},\\
    \psi =& \arccos(\theta_1/r).
\end{align*}
The tail probabilities are then obtained via numerical integration of $\rho$ using \texttt{mpmath}.

The second class is adopted from \citep{dai2014multi}, which admits a product-form Laplace transform due to the {\em skew-symmetry} property. An RBM is said to satisfy the skew-symmetry condition if
\begin{align*}
    2\Sigma = R {\rm diag}(R)^{-1} {\rm diag}(\Sigma) + {\rm diag}(\Sigma) {\rm diag}(R)^{-1} R^T.
\end{align*}
Following \citep{dai2014multi}, we construct an RBM for any dimension $d \geq 2$ that satisfies the skew symmetry condition as follows:
\begin{align*}
    &R_{j,j} = 1, && \forall j = 1, \dots, d,\\
    &R_{j,j-1} = -1, && \forall j = 2, \dots, d,\\
    &\Sigma_{j,j} = c_j + c_{j+1}, && \forall j = 1, \dots, d,\\
    &\Sigma_{j,j-1} = -c_j, && \forall j = 2, \dots, d,\\
    &\Sigma_{j-1, j} = -c_j, && \forall j = 2, \dots, d,\\
    &\mu_j = \beta_j - \beta_{j+1}, && \forall j = 1, \dots, d,
\end{align*}
where the vectors $c \in \mathbb{R}^{d+1}$ and $\beta \in \mathbb{R}^{d+1}$ are defined by
\begin{align*}
    c_j =& 1, && \forall j = 1, \dots, d + 1,\\
    \beta_j =& j, && \forall j = 1, \dots, d + 1.
\end{align*}
According to \cite{dai2014multi}, the corresponding Laplace transforms are given by
\begin{align*}
    \varphi_0(\theta) =& \prod_{j = 1}^d \frac{\alpha_j}{\alpha_j - \theta_j}, && \forall \theta < \alpha,\\
    \varphi_k(\theta) =& \frac{\Sigma_{k,k}}{2R_{k,k}} \alpha_k \prod_{j \neq k} \frac{\alpha_j}{\alpha_j - \theta_j}, && \forall \theta < \alpha,
\end{align*}
where $\alpha$ is defined as
\begin{align*}
    \alpha = -2 {\rm diag}(\Sigma)^{-1} {\rm diag}(R) R^{-1} \mu.
\end{align*}
The tail probabilities are then computed via numerical Laplace inversion using the Talbot method implemented in \texttt{mpmath}.

\section{Moments Estimation}
\subsection{Numerical procedure for computing moments given Laplace transform}
Given the Laplace transform, we compute the moments of the RBM stationary distribution by following the procedure in \cite{choudhury1996numerical}. To make this paper self-contained, we summarize the key steps below.

The method is based on numerical inversion of the Laplace transform in the complex plane. For each moment order $n$, the inversion is carried out along a circular contour 
\[
    \mathcal{C}_n := \{ z \in \mathbb{C} : |z| = r_n \},
\]
centered at the origin with radius $r_n$. The contour integral representation of the moment is then approximated by evaluating the integrand at equally spaced points 
\[
    z_j = r_n e^{i\pi j/(n\ell)}, \quad j = 0,1,\dots,n\ell,
\]
which leads to a trapezoidal-type discretization of the integral. This yields the following inversion formula:
\begin{align} \label{eq:moment-inversion}
    m_n = \frac{n!}{2n\ell\,r_n^n\,a_n^n} \left[ W_n(r_n)+(-1)^nW_n(-r_n) + 2\sum_{j=1}^{n\ell-1} \rm Re\! \left(W_n\!(z_j)e^{-i\pi j}\right) \right],
\end{align}
where $r_n = 10^{-\,\epsilon/(2n \ell)}$ for some accuracy parameter $\epsilon > 0$ and inversion parameter $\ell \in \{1, 2\}$. The parameter $r_n$ is chosen to control the discretization error arising from approximating the contour integral using a finite trapezoidal sum, while the parameter $\ell$ helps mitigate round-off error by increasing the number of quadrature points along the contour. The function $W_n(z) := \varphi(-a_n z)$ is a rescaled version of the Laplace transform. The scaling factors $\{a_n\}$ are chosen adaptively so that the magnitude of $W_n(z)$ remains well-conditioned on the contour $\mathcal{C}_n$, preventing numerical overflow or underflow when computing high-order moments.

The overall procedure is summarized in Algorithm~\ref{alg:adaptive-moment-laplace}. It first computes the first two moments, refines them using improved scaling, and then proceeds recursively to higher-order moments using previously computed values to update the scaling factors.

\begin{algorithm}[t]
    \small
    \caption{Adaptive moment computation from Laplace transform}
    \label{alg:adaptive-moment-laplace}
    \DontPrintSemicolon
    \SetKwInOut{Input}{Input}
    \SetKwInOut{Output}{Output}
    
    \Input{Laplace transform $\varphi(\theta)$, number of moments $M\ge 2$, accuracy parameter $\epsilon$, inversion parameter $\ell \in\{1,2\}$}
    
    \BlankLine
    // Step 1: Initial computation of $m_1$ and $m_2$\;
    Set $a_1 \leftarrow 1$ and $a_2 \leftarrow m_1$\;
    Compute $m_1$ and $m_2$ using \eqref{eq:moment-inversion} with $W_1(z)=\varphi(-a_1 z)$ and $W_2(z)=\varphi(-a_2 z)$\;
    
    \BlankLine
    // Step 2: Refinement of $m_1$ and $m_2$\;
    Set $a_1 \leftarrow \frac{2m_1}{m_2}$ and $a_2 \leftarrow \frac{2m_1}{m_2}$\;
    Recompute $m_1,m_2$ using \eqref{eq:moment-inversion} with $W_1(z)=\varphi(-a_1 z)$ and $W_2(z)=\varphi(-a_2 z)$\;
    
    \BlankLine
    // Step 3: Compute $m_n$ for $3 \leq n \leq M$\;
    \For{$n=3,4,\dots,M$}{
        Set $a_n \leftarrow \frac{(n-1)m_{n-2}}{m_{n-1}}$\;
        Compute $m_n$ using \eqref{eq:moment-inversion} with $W_n(z) = \varphi(-a_n z)$\;
    }
    
    \Return $m_1,m_2,\dots,m_M$\;
\end{algorithm}

\subsection{Experiment results}
Using the procedure described in Algorithm \ref{alg:adaptive-moment-laplace}, we estimate the moments of RBMs adopted from \cite{dai2014multi} with dimensions $5$, $20$, and $30$. The results are reported in Tables \ref{tab:moments-dim-5}--\ref{tab:moments-dim-30}. Overall, the proposed framework provides accurate estimates for low-order moments, particularly in lower-dimensional settings. For the $5$-dimensional RBM, the estimated first-, second-, and third-order moments all exhibit relatively small relative errors. For the $20$- and $30$-dimensional RBMs, the first-order moments also remain reasonably accurate across most coordinates. As the dimension and moment order increase, however, the estimation errors become larger, especially for certain second-order moments in the $30$-dimensional case. This behavior is expected, as moment estimation relies on accurate local evaluations of the Laplace transform near $\theta = 0$ through contour-based derivative approximation, which becomes increasingly sensitive to approximation errors in high-dimensional settings. In contrast, the Talbot inversion used for tail probability estimation tends to remain more stable in high dimensions because it depends on evaluating the Laplace transform over a broader region of the complex plane rather than accurately recovering local derivative information near $\theta = 0$. As a result, small local approximation errors are less significantly amplified compared with moment estimation.

\begin{table}[htb]
    \small
    \centering
    \begin{tabular}{c c c c c}
    \toprule
    \multicolumn{5}{c}{\textbf{First Moments}} \\
    \midrule
    Dimension $j$ & True & Pred & Abs Err & Rel Err \\
    \midrule
    0 & $1.00 \times 10^{0}$ & $9.98 \times 10^{-1}$ & $2.06 \times 10^{-3}$ & $2.06 \times 10^{-3}$ \\
    1 & $5.00 \times 10^{-1}$ & $4.99 \times 10^{-1}$ & $1.19 \times 10^{-3}$ & $2.38 \times 10^{-3}$ \\
    2 & $3.33 \times 10^{-1}$ & $3.34 \times 10^{-1}$ & $6.46 \times 10^{-4}$ & $1.94 \times 10^{-3}$ \\
    3 & $2.50 \times 10^{-1}$ & $2.51 \times 10^{-1}$ & $6.83 \times 10^{-4}$ & $2.73 \times 10^{-3}$ \\
    4 & $2.00 \times 10^{-1}$ & $2.00 \times 10^{-1}$ & $5.72 \times 10^{-5}$ & $2.86 \times 10^{-4}$ \\
    \midrule
    \multicolumn{5}{c}{\textbf{Second Moments}} \\
    % \midrule
    % Dimension $i$ & True & Pred & Abs Err & Rel Err \\
    \midrule
    0 & $2.00 \times 10^{0}$ & $2.00 \times 10^{0}$ & $3.09 \times 10^{-3}$ & $1.55 \times 10^{-3}$ \\
    1 & $5.00 \times 10^{-1}$ & $4.99 \times 10^{-1}$ & $8.76 \times 10^{-4}$ & $1.75 \times 10^{-3}$ \\
    2 & $2.22 \times 10^{-1}$ & $2.25 \times 10^{-1}$ & $2.28 \times 10^{-3}$ & $1.03 \times 10^{-2}$ \\
    3 & $1.25 \times 10^{-1}$ & $1.27 \times 10^{-1}$ & $2.18 \times 10^{-3}$ & $1.74 \times 10^{-2}$ \\
    4 & $8.00 \times 10^{-2}$ & $7.85 \times 10^{-2}$ & $1.45 \times 10^{-3}$ & $1.82 \times 10^{-2}$ \\
    \midrule
    \multicolumn{5}{c}{\textbf{Third Moments}} \\
    % \midrule
    % Dimension $i$ & True & Pred & Abs Err & Rel Err \\
    \midrule
    0 & $6.00 \times 10^{0}$ & $5.88 \times 10^{0}$ & $1.16 \times 10^{-1}$ & $1.93 \times 10^{-2}$ \\
    1 & $7.50 \times 10^{-1}$ & $7.27 \times 10^{-1}$ & $2.31 \times 10^{-2}$ & $3.08 \times 10^{-2}$ \\
    2 & $2.22 \times 10^{-1}$ & $2.27 \times 10^{-1}$ & $4.71 \times 10^{-3}$ & $2.12 \times 10^{-2}$ \\
    3 & $9.38 \times 10^{-2}$ & $9.60 \times 10^{-2}$ & $2.22 \times 10^{-3}$ & $2.36 \times 10^{-2}$ \\
    4 & $4.80 \times 10^{-2}$ & $4.32 \times 10^{-2}$ & $4.84 \times 10^{-3}$ & $1.01 \times 10^{-1}$ \\
    \bottomrule
    \end{tabular}
    \caption{Moment estimates for $d=5$}
    \label{tab:moments-dim-5}
\end{table}

\begin{table}[htb]
    \small
    \centering
    \begin{tabular}{c c c c c}
    \toprule
    \multicolumn{5}{c}{\textbf{First Moments}} \\
    \midrule
    Dimension $j$ & True & Pred & Abs Err & Rel Err \\
    \midrule
    0 & $1.00 \times 10^{0}$ & $9.94 \times 10^{-1}$ & $5.84 \times 10^{-3}$ & $5.84 \times 10^{-3}$ \\
    1 & $5.00 \times 10^{-1}$ & $4.89 \times 10^{-1}$ & $1.10 \times 10^{-2}$ & $2.20 \times 10^{-2}$ \\
    2 & $3.33 \times 10^{-1}$ & $3.28 \times 10^{-1}$ & $5.17 \times 10^{-3}$ & $1.55 \times 10^{-2}$ \\
    3 & $2.50 \times 10^{-1}$ & $2.47 \times 10^{-1}$ & $2.55 \times 10^{-3}$ & $1.02 \times 10^{-2}$ \\
    4 & $2.00 \times 10^{-1}$ & $1.99 \times 10^{-1}$ & $1.45 \times 10^{-3}$ & $7.27 \times 10^{-3}$ \\
    5 & $1.67 \times 10^{-1}$ & $1.64 \times 10^{-1}$ & $2.49 \times 10^{-3}$ & $1.49 \times 10^{-2}$ \\
    6 & $1.43 \times 10^{-1}$ & $1.39 \times 10^{-1}$ & $3.77 \times 10^{-3}$ & $2.64 \times 10^{-2}$ \\
    7 & $1.25 \times 10^{-1}$ & $1.23 \times 10^{-1}$ & $1.70 \times 10^{-3}$ & $1.36 \times 10^{-2}$ \\
    8 & $1.11 \times 10^{-1}$ & $1.09 \times 10^{-1}$ & $2.61 \times 10^{-3}$ & $2.35 \times 10^{-2}$ \\
    9 & $1.00 \times 10^{-1}$ & $9.70 \times 10^{-2}$ & $3.01 \times 10^{-3}$ & $3.01 \times 10^{-2}$ \\
    10 & $9.09 \times 10^{-2}$ & $8.82 \times 10^{-2}$ & $2.74 \times 10^{-3}$ & $3.02 \times 10^{-2}$ \\
    11 & $8.33 \times 10^{-2}$ & $8.07 \times 10^{-2}$ & $2.65 \times 10^{-3}$ & $3.18 \times 10^{-2}$ \\
    12 & $7.69 \times 10^{-2}$ & $7.46 \times 10^{-2}$ & $2.36 \times 10^{-3}$ & $3.07 \times 10^{-2}$ \\
    13 & $7.14 \times 10^{-2}$ & $6.90 \times 10^{-2}$ & $2.41 \times 10^{-3}$ & $3.38 \times 10^{-2}$ \\
    14 & $6.67 \times 10^{-2}$ & $6.49 \times 10^{-2}$ & $1.77 \times 10^{-3}$ & $2.65 \times 10^{-2}$ \\
    15 & $6.25 \times 10^{-2}$ & $6.07 \times 10^{-2}$ & $1.83 \times 10^{-3}$ & $2.94 \times 10^{-2}$ \\
    16 & $5.88 \times 10^{-2}$ & $5.65 \times 10^{-2}$ & $2.30 \times 10^{-3}$ & $3.90 \times 10^{-2}$ \\
    17 & $5.56 \times 10^{-2}$ & $5.32 \times 10^{-2}$ & $2.31 \times 10^{-3}$ & $4.15 \times 10^{-2}$ \\
    18 & $5.26 \times 10^{-2}$ & $5.00 \times 10^{-2}$ & $2.58 \times 10^{-3}$ & $4.91 \times 10^{-2}$ \\
    19 & $5.00 \times 10^{-2}$ & $4.82 \times 10^{-2}$ & $1.80 \times 10^{-3}$ & $3.60 \times 10^{-2}$ \\
    \midrule
    \multicolumn{5}{c}{\textbf{Second Moments}} \\
    % \midrule
    % Dimension $i$ & True & Pred & Abs Err & Rel Err \\
    \midrule
    0 & $2.00 \times 10^{0}$ & $1.95 \times 10^{0}$ & $5.37 \times 10^{-2}$ & $2.69 \times 10^{-2}$ \\
    1 & $5.00 \times 10^{-1}$ & $4.75 \times 10^{-1}$ & $2.45 \times 10^{-2}$ & $4.91 \times 10^{-2}$ \\
    2 & $2.22 \times 10^{-1}$ & $2.23 \times 10^{-1}$ & $1.14 \times 10^{-3}$ & $5.12 \times 10^{-3}$ \\
    3 & $1.25 \times 10^{-1}$ & $1.23 \times 10^{-1}$ & $1.83 \times 10^{-3}$ & $1.47 \times 10^{-2}$ \\
    4 & $8.00 \times 10^{-2}$ & $7.91 \times 10^{-2}$ & $9.47 \times 10^{-4}$ & $1.18 \times 10^{-2}$ \\
    5 & $5.56 \times 10^{-2}$ & $5.40 \times 10^{-2}$ & $1.55 \times 10^{-3}$ & $2.79 \times 10^{-2}$ \\
    6 & $4.08 \times 10^{-2}$ & $3.80 \times 10^{-2}$ & $2.81 \times 10^{-3}$ & $6.90 \times 10^{-2}$ \\
    7 & $3.13 \times 10^{-2}$ & $3.00 \times 10^{-2}$ & $1.29 \times 10^{-3}$ & $4.13 \times 10^{-2}$ \\
    8 & $2.47 \times 10^{-2}$ & $2.30 \times 10^{-2}$ & $1.73 \times 10^{-3}$ & $7.02 \times 10^{-2}$ \\
    9 & $2.00 \times 10^{-2}$ & $1.84 \times 10^{-2}$ & $1.65 \times 10^{-3}$ & $8.25 \times 10^{-2}$ \\
    10 & $1.65 \times 10^{-2}$ & $1.48 \times 10^{-2}$ & $1.72 \times 10^{-3}$ & $1.04 \times 10^{-1}$ \\
    11 & $1.39 \times 10^{-2}$ & $5.44 \times 10^{-3}$ & $8.45 \times 10^{-3}$ & $6.09 \times 10^{-1}$ \\
    12 & $1.18 \times 10^{-2}$ & $1.07 \times 10^{-2}$ & $1.15 \times 10^{-3}$ & $9.74 \times 10^{-2}$ \\
    13 & $1.02 \times 10^{-2}$ & $7.04 \times 10^{-3}$ & $3.17 \times 10^{-3}$ & $3.11 \times 10^{-1}$ \\
    14 & $8.89 \times 10^{-3}$ & $8.25 \times 10^{-3}$ & $6.36 \times 10^{-4}$ & $7.16 \times 10^{-2}$ \\
    15 & $7.81 \times 10^{-3}$ & $4.83 \times 10^{-3}$ & $2.98 \times 10^{-3}$ & $3.82 \times 10^{-1}$ \\
    16 & $6.92 \times 10^{-3}$ & $6.09 \times 10^{-3}$ & $8.32 \times 10^{-4}$ & $1.20 \times 10^{-1}$ \\
    17 & $6.17 \times 10^{-3}$ & $5.21 \times 10^{-3}$ & $9.66 \times 10^{-4}$ & $1.56 \times 10^{-1}$ \\
    18 & $5.54 \times 10^{-3}$ & $3.79 \times 10^{-3}$ & $1.75 \times 10^{-3}$ & $3.15 \times 10^{-1}$ \\
    19 & $5.00 \times 10^{-3}$ & $3.65 \times 10^{-3}$ & $1.35 \times 10^{-3}$ & $2.70 \times 10^{-1}$ \\
    \bottomrule
    \end{tabular}
    \caption{Moment estimates for $d=20$}
    \label{tab:moments-dim-20}
\end{table}

\begin{center}
\small
\begin{longtable}{c c c c c}
    \toprule
    \multicolumn{5}{c}{\textbf{First Moments}} \\
    \midrule
    Dimension $j$ & True & Pred & Abs Err & Rel Err \\
    \midrule
    \endfirsthead
    
    \toprule
    Dimension $j$ & True & Pred & Abs Err & Rel Err \\
    \midrule
    \endhead
    
    \bottomrule
    \endfoot
    
    0 & $1.00 \times 10^{0}$ & $9.90 \times 10^{-1}$ & $1.04 \times 10^{-2}$ & $1.04 \times 10^{-2}$ \\
    1 & $5.00 \times 10^{-1}$ & $4.75 \times 10^{-1}$ & $2.48 \times 10^{-2}$ & $4.97 \times 10^{-2}$ \\
    2 & $3.33 \times 10^{-1}$ & $3.19 \times 10^{-1}$ & $1.47 \times 10^{-2}$ & $4.42 \times 10^{-2}$ \\
    3 & $2.50 \times 10^{-1}$ & $2.37 \times 10^{-1}$ & $1.32 \times 10^{-2}$ & $5.28 \times 10^{-2}$ \\
    4 & $2.00 \times 10^{-1}$ & $1.88 \times 10^{-1}$ & $1.16 \times 10^{-2}$ & $5.81 \times 10^{-2}$ \\
    5 & $1.67 \times 10^{-1}$ & $1.53 \times 10^{-1}$ & $1.34 \times 10^{-2}$ & $8.02 \times 10^{-2}$ \\
    6 & $1.43 \times 10^{-1}$ & $1.32 \times 10^{-1}$ & $1.07 \times 10^{-2}$ & $7.52 \times 10^{-2}$ \\
    7 & $1.25 \times 10^{-1}$ & $1.17 \times 10^{-1}$ & $8.06 \times 10^{-3}$ & $6.45 \times 10^{-2}$ \\
    8 & $1.11 \times 10^{-1}$ & $1.04 \times 10^{-1}$ & $6.97 \times 10^{-3}$ & $6.27 \times 10^{-2}$ \\
    9 & $1.00 \times 10^{-1}$ & $9.29 \times 10^{-2}$ & $7.11 \times 10^{-3}$ & $7.11 \times 10^{-2}$ \\
    10 & $9.09 \times 10^{-2}$ & $8.54 \times 10^{-2}$ & $5.48 \times 10^{-3}$ & $6.03 \times 10^{-2}$ \\
    11 & $8.33 \times 10^{-2}$ & $7.79 \times 10^{-2}$ & $5.42 \times 10^{-3}$ & $6.51 \times 10^{-2}$ \\
    12 & $7.69 \times 10^{-2}$ & $7.22 \times 10^{-2}$ & $4.69 \times 10^{-3}$ & $6.09 \times 10^{-2}$ \\
    13 & $7.14 \times 10^{-2}$ & $6.72 \times 10^{-2}$ & $4.27 \times 10^{-3}$ & $5.98 \times 10^{-2}$ \\
    14 & $6.67 \times 10^{-2}$ & $6.23 \times 10^{-2}$ & $4.34 \times 10^{-3}$ & $6.51 \times 10^{-2}$ \\
    15 & $6.25 \times 10^{-2}$ & $5.87 \times 10^{-2}$ & $3.77 \times 10^{-3}$ & $6.03 \times 10^{-2}$ \\
    16 & $5.88 \times 10^{-2}$ & $5.58 \times 10^{-2}$ & $3.06 \times 10^{-3}$ & $5.20 \times 10^{-2}$ \\
    17 & $5.56 \times 10^{-2}$ & $5.32 \times 10^{-2}$ & $2.31 \times 10^{-3}$ & $4.16 \times 10^{-2}$ \\
    18 & $5.26 \times 10^{-2}$ & $5.07 \times 10^{-2}$ & $1.91 \times 10^{-3}$ & $3.64 \times 10^{-2}$ \\
    19 & $5.00 \times 10^{-2}$ & $4.82 \times 10^{-2}$ & $1.81 \times 10^{-3}$ & $3.62 \times 10^{-2}$ \\
    20 & $4.76 \times 10^{-2}$ & $4.60 \times 10^{-2}$ & $1.66 \times 10^{-3}$ & $3.49 \times 10^{-2}$ \\
    21 & $4.55 \times 10^{-2}$ & $4.42 \times 10^{-2}$ & $1.22 \times 10^{-3}$ & $2.67 \times 10^{-2}$ \\
    22 & $4.35 \times 10^{-2}$ & $4.26 \times 10^{-2}$ & $9.28 \times 10^{-4}$ & $2.13 \times 10^{-2}$ \\
    23 & $4.17 \times 10^{-2}$ & $4.06 \times 10^{-2}$ & $1.11 \times 10^{-3}$ & $2.67 \times 10^{-2}$ \\
    24 & $4.00 \times 10^{-2}$ & $3.89 \times 10^{-2}$ & $1.07 \times 10^{-3}$ & $2.68 \times 10^{-2}$ \\
    25 & $3.85 \times 10^{-2}$ & $3.74 \times 10^{-2}$ & $1.07 \times 10^{-3}$ & $2.78 \times 10^{-2}$ \\
    26 & $3.70 \times 10^{-2}$ & $3.60 \times 10^{-2}$ & $1.00 \times 10^{-3}$ & $2.71 \times 10^{-2}$ \\
    27 & $3.57 \times 10^{-2}$ & $3.51 \times 10^{-2}$ & $6.14 \times 10^{-4}$ & $1.72 \times 10^{-2}$ \\
    28 & $3.45 \times 10^{-2}$ & $3.34 \times 10^{-2}$ & $1.11 \times 10^{-3}$ & $3.23 \times 10^{-2}$ \\
    29 & $3.33 \times 10^{-2}$ & $3.23 \times 10^{-2}$ & $1.03 \times 10^{-3}$ & $3.08 \times 10^{-2}$ \\
    
    \midrule
    \multicolumn{5}{c}{\textbf{Second Moments}} \\
    \midrule
    
    0 & $2.00 \times 10^{0}$ & $1.91 \times 10^{0}$ & $8.74 \times 10^{-2}$ & $4.37 \times 10^{-2}$ \\
    1 & $5.00 \times 10^{-1}$ & $4.50 \times 10^{-1}$ & $4.97 \times 10^{-2}$ & $9.93 \times 10^{-2}$ \\
    2 & $2.22 \times 10^{-1}$ & $2.07 \times 10^{-1}$ & $1.50 \times 10^{-2}$ & $6.77 \times 10^{-2}$ \\
    3 & $1.25 \times 10^{-1}$ & $1.18 \times 10^{-1}$ & $6.76 \times 10^{-3}$ & $5.41 \times 10^{-2}$ \\
    4 & $8.00 \times 10^{-2}$ & $7.47 \times 10^{-2}$ & $5.33 \times 10^{-3}$ & $6.67 \times 10^{-2}$ \\
    5 & $5.56 \times 10^{-2}$ & $4.86 \times 10^{-2}$ & $6.97 \times 10^{-3}$ & $1.25 \times 10^{-1}$ \\
    6 & $4.08 \times 10^{-2}$ & $3.44 \times 10^{-2}$ & $6.39 \times 10^{-3}$ & $1.57 \times 10^{-1}$ \\
    7 & $3.13 \times 10^{-2}$ & $1.85 \times 10^{-2}$ & $1.28 \times 10^{-2}$ & $4.09 \times 10^{-1}$ \\
    8 & $2.47 \times 10^{-2}$ & $2.07 \times 10^{-2}$ & $4.04 \times 10^{-3}$ & $1.64 \times 10^{-1}$ \\
    9 & $2.00 \times 10^{-2}$ & $1.26 \times 10^{-2}$ & $7.42 \times 10^{-3}$ & $3.71 \times 10^{-1}$ \\
    10 & $1.65 \times 10^{-2}$ & $9.04 \times 10^{-3}$ & $7.49 \times 10^{-3}$ & $4.53 \times 10^{-1}$ \\
    11 & $1.39 \times 10^{-2}$ & $8.66 \times 10^{-3}$ & $5.23 \times 10^{-3}$ & $3.77 \times 10^{-1}$ \\
    12 & $1.18 \times 10^{-2}$ & $-2.12 \times 10^{-3}$ & $1.40 \times 10^{-2}$ & $1.18 \times 10^{0}$ \\
    13 & $1.02 \times 10^{-2}$ & $2.56 \times 10^{-3}$ & $7.65 \times 10^{-3}$ & $7.49 \times 10^{-1}$ \\
    14 & $8.89 \times 10^{-3}$ & $2.26 \times 10^{-3}$ & $6.63 \times 10^{-3}$ & $7.45 \times 10^{-1}$ \\
    15 & $7.81 \times 10^{-3}$ & $6.45 \times 10^{-3}$ & $1.37 \times 10^{-3}$ & $1.75 \times 10^{-1}$ \\
    16 & $6.92 \times 10^{-3}$ & $5.53 \times 10^{-3}$ & $1.39 \times 10^{-3}$ & $2.01 \times 10^{-1}$ \\
    17 & $6.17 \times 10^{-3}$ & $3.53 \times 10^{-3}$ & $2.64 \times 10^{-3}$ & $4.28 \times 10^{-1}$ \\
    18 & $5.54 \times 10^{-3}$ & $2.21 \times 10^{-3}$ & $3.33 \times 10^{-3}$ & $6.01 \times 10^{-1}$ \\
    19 & $5.00 \times 10^{-3}$ & $4.34 \times 10^{-3}$ & $6.56 \times 10^{-4}$ & $1.31 \times 10^{-1}$ \\
    20 & $4.54 \times 10^{-3}$ & $-3.56 \times 10^{-3}$ & $8.09 \times 10^{-3}$ & $1.78 \times 10^{0}$ \\
    21 & $4.13 \times 10^{-3}$ & $-3.49 \times 10^{-4}$ & $4.48 \times 10^{-3}$ & $1.08 \times 10^{0}$ \\
    22 & $3.78 \times 10^{-3}$ & $-6.16 \times 10^{-4}$ & $4.40 \times 10^{-3}$ & $1.16 \times 10^{0}$ \\
    23 & $3.47 \times 10^{-3}$ & $9.72 \times 10^{-5}$ & $3.38 \times 10^{-3}$ & $9.72 \times 10^{-1}$ \\
    24 & $3.20 \times 10^{-3}$ & $-2.92 \times 10^{-3}$ & $6.12 \times 10^{-3}$ & $1.91 \times 10^{0}$ \\
    25 & $2.96 \times 10^{-3}$ & $-2.83 \times 10^{-3}$ & $5.79 \times 10^{-3}$ & $1.96 \times 10^{0}$ \\
    26 & $2.74 \times 10^{-3}$ & $-3.85 \times 10^{-3}$ & $6.60 \times 10^{-3}$ & $2.40 \times 10^{0}$ \\
    27 & $2.55 \times 10^{-3}$ & $8.05 \times 10^{-5}$ & $2.47 \times 10^{-3}$ & $9.68 \times 10^{-1}$ \\
    28 & $2.38 \times 10^{-3}$ & $1.71 \times 10^{-3}$ & $6.69 \times 10^{-4}$ & $2.81 \times 10^{-1}$ \\
    29 & $2.22 \times 10^{-3}$ & $1.58 \times 10^{-3}$ & $6.43 \times 10^{-4}$ & $2.89 \times 10^{-1}$ \\
    
    \caption{Moment estimates for $d=30$}
    \label{tab:moments-dim-30}
    
\end{longtable}
\end{center}

\clearpage

% \ACKNOWLEDGMENT{%
% % Enter the text of acknowledgments here
% }% Leave this (end of acknowledgment)

% \section{DRAFT}
% \input{draft}

% Acknowledgments here

% Appendix here
% Options are (1) APPENDIX (with or without general title) or 
%             (2) APPENDICES (if it has more than one unrelated sections)
% Outcomment the appropriate case if necessary
%
% \begin{APPENDIX}{<Title of the Appendix>}
% \end{APPENDIX}
%
%   or 
%
% \begin{APPENDICES}
% \section{<Title of Section A>}
% \section{<Title of Section B>}
% etc
% \end{APPENDICES}

% References here (outcomment the appropriate case) 

% CASE 1: BiBTeX used to constantly update the references 
%   (while the paper is being written).
\bibliographystyle{unsrtnat} % outcomment this and next line in Case 1
\bibliography{references} % if more than one, comma separated

\end{document}